\documentclass[11pt]{article}

\usepackage[preprint]{acl}

\usepackage{times}
\usepackage{latexsym}
\usepackage[T1]{fontenc}
\usepackage[utf8]{inputenc}
\usepackage{microtype}
\usepackage{inconsolata}
\usepackage{graphicx}
\usepackage{booktabs}
\usepackage{multirow}
\usepackage{array}
\usepackage{xcolor}
\usepackage{enumitem}
\usepackage{amsmath}
\usepackage{amssymb}
\usepackage{xurl}
\usepackage{placeins}
\usepackage{float}
\usepackage[most]{tcolorbox}
\title{PerspectiveGap: A Benchmark for Multi-Agent Orchestration Prompting}

\author{Youran Sun$^{1,*}$ \quad Xingyu Ren$^{2,*}$ \quad Kejia Zhang$^1$ \quad Xinpeng Liu \quad Jiaxuan Guo$^{3,\dagger}$}

\begin{document}
\maketitle
\begingroup
\renewcommand{\thefootnote}{}
\footnotetext{$^1$University of Maryland. $^2$The Chinese University of Hong Kong. $^3$Stanford University. $^*$Equal contribution. $^\dagger$Corresponding author. Emails: Youran Sun, \texttt{sun1245@umd.edu}; Jiaxuan Guo, \texttt{guojx@stanford.edu}.}
\endgroup

\begin{abstract}
Real-world LLM applications are moving beyond single-agent workflows toward orchestrated multi-agent systems, yet current models still struggle to determine what each sub-agent needs to know.
To measure this, we introduce PerspectiveGap, a benchmark for evaluating LLMs' ability to compose orchestration prompts for multi-agent systems.
PerspectiveGap contains 110 scenarios, each evaluated through two distractor-mixed task formats: role-fragment assignment and free-form prompt writing.
These scenarios are organized into 10 topologies, which are distilled from the authors' real-world engineering practice and framed by the Prompt Economy principle: building loop-centered orchestrations that maximize utility with minimal role and engineering overhead.
In experiments with 33 commercial models from 10 companies, GPT-5.5 substantially outperforms all competitors, whereas Opus 4.8 shows a notable weakness in orchestration prompting despite its strong coding performance.
Nevertheless, PerspectiveGap remains challenging: the evaluated models achieve an average combined pass rate of only 17.2\% (GPT-5.5 62.0\%) and an average overall leakage rate of 217.9\% (a per-scenario information leak-event count, not a proportion; GPT-5.5 49.1\%).
These findings suggest that multi-agent orchestration prompting is a distinct and under-evaluated capability, and PerspectiveGap provides a foundation for measuring and improving it systematically.
\end{abstract}

\section{Introduction}
\label{sec:intro}

Prompt engineering has shifted from tuning a single prompt to designing multi-agent orchestras, in which a task is decomposed into specialized roles with distinct information boundaries and artifact handoffs \citep{li2023camel,chatdev2023,metagpt2023,wu2023autogen,chen2023agentverse,sakana2024aiscientist,tran2025multiagent}.

Constructing these orchestras requires orchestration prompting: writing sub-agent instructions that specify each role's task scope, context boundaries, and expected handoffs.
Yet current LLMs struggle to determine what each sub-agent needs to know.
The resulting failures are not cosmetic: main agents leak distractors, expose out-of-role information, drop shared context, confuse artifact ownership, and sometimes place instructions where the sub-agent cannot see them.
These errors produce incomplete, contaminated, or self-defeating sub-agent prompts.
Section~\ref{sec:analysis} analyzes these failure modes in detail.

Existing evaluations do not target this orchestration-prompting ability.
Theory-of-mind (ToM) benchmarks typically score question answering or belief tracking \citep{le2019tomi,kim2023fantom,sclar2024exploretom}, multiple-choice action prediction \citep{gu2024simpletom}, dialogue acts \citep{ys2026sotopiatom}, or functional behavior labels \citep{riemer2025tombroken}.
Agent benchmarks instead score downstream task success, tool use, environment-level performance, or instruction following \citep{liu2023agentbench,qin2023toolllm,orogat2026understanding,qi2025agentif}.
Neither line directly evaluates whether a main agent can write sub-agent prompts that respect asymmetric context and role-specific information needs.
Section~\ref{sec:related} discusses this boundary.

\textbf{We introduce PerspectiveGap, to our knowledge the first benchmark for multi-agent orchestration prompt writing.}
PerspectiveGap contains 110 scenarios, each consisting of a role list, a shuffled set of information fragments $f_1,\ldots,f_N$, and a reference answer specifying which fragments each role needs.
Each scenario includes two tasks: role-fragment assignment and free-form prompt writing.
The former asks the model to output the fragment IDs for each role; the latter asks the model to write the actual sub-agent prompts.
The two tasks share the same fragments, so comparing them isolates models that can identify each role's needs but fail to act on them when writing the prompt.
Each scenario also injects one distractor, such as prompt-engineering advice, that may look useful to the tested LLM but is useless or even burdensome for any sub-agent.
The benchmark covers orchestrations with 2--6 roles and 7--13 information fragments.
A deterministic, rule-only scorer evaluates each role's prompt for inclusion of required fragments and exclusion of irrelevant content; a 716-row hand audit validates this containment detector.
The construction pipeline records each scenario's topology, role definitions, fragments, and reference answers explicitly, allowing new scenarios to be added with auditable answer keys.

The 110 scenarios are organized into 10 topologies shown in Figure~\ref{fig:patterns}.
These topologies are distilled from the authors' real-world prompt-engineering practice and from production agent-system patterns.
We frame their loop-centered structure with Prompt Economy.
Under this framing, engineering cost is largely fixed by the number of role prompts, while benefit accumulates across repeated role invocations.
Loop-centered orchestration patterns are therefore attractive when the goal is to maximize utility with minimal engineering overhead.
The release uses six base patterns plus four pool variants, all built around one or more critic loops.
As a set, they cover the main practical one- and two-loop orchestration patterns targeted by PerspectiveGap.

We evaluate 33 commercial models from 10 companies on PerspectiveGap.
GPT-5.5 leads with a 62.0\% combined pass rate; gpt-5.6-terra and gpt-5.6-sol follow at 42.7\% and 35.7\%, and models ranked 4--9 cluster between 22\% and 32\%.
The average combined pass rate is 17.2\%.
The average overall leakage rate is 217.9\%, compared with 49.1\% for GPT-5.5.
Opus 4.8 underperforms relative to its strong coding performance~\citep{wang2025swe}\footnote{This finding also matches the authors' day-to-day experience; PerspectiveGap finally quantifies that intuition.}, suggesting that orchestration prompting is not reducible to coding skill.
These results indicate that multi-agent orchestration prompting is a distinct and under-evaluated capability.

This paper makes four contributions:
\begin{enumerate}
    \item We release PerspectiveGap, a 110-scenario benchmark with two task formats per scenario, per-scenario answer keys, construction notes, and full logs.
    \item We evaluate 33 commercial models from 10 companies and report a standardized leaderboard, revealing low average performance.
    \item We identify five recurring failure modes in multi-agent orchestration prompting.
    \item We articulate \textbf{Prompt Economy} as a cost-benefit framing for the loop-centered topology family.
\end{enumerate}

\begin{figure}[!htbp]
\centering
\includegraphics[width=\columnwidth]{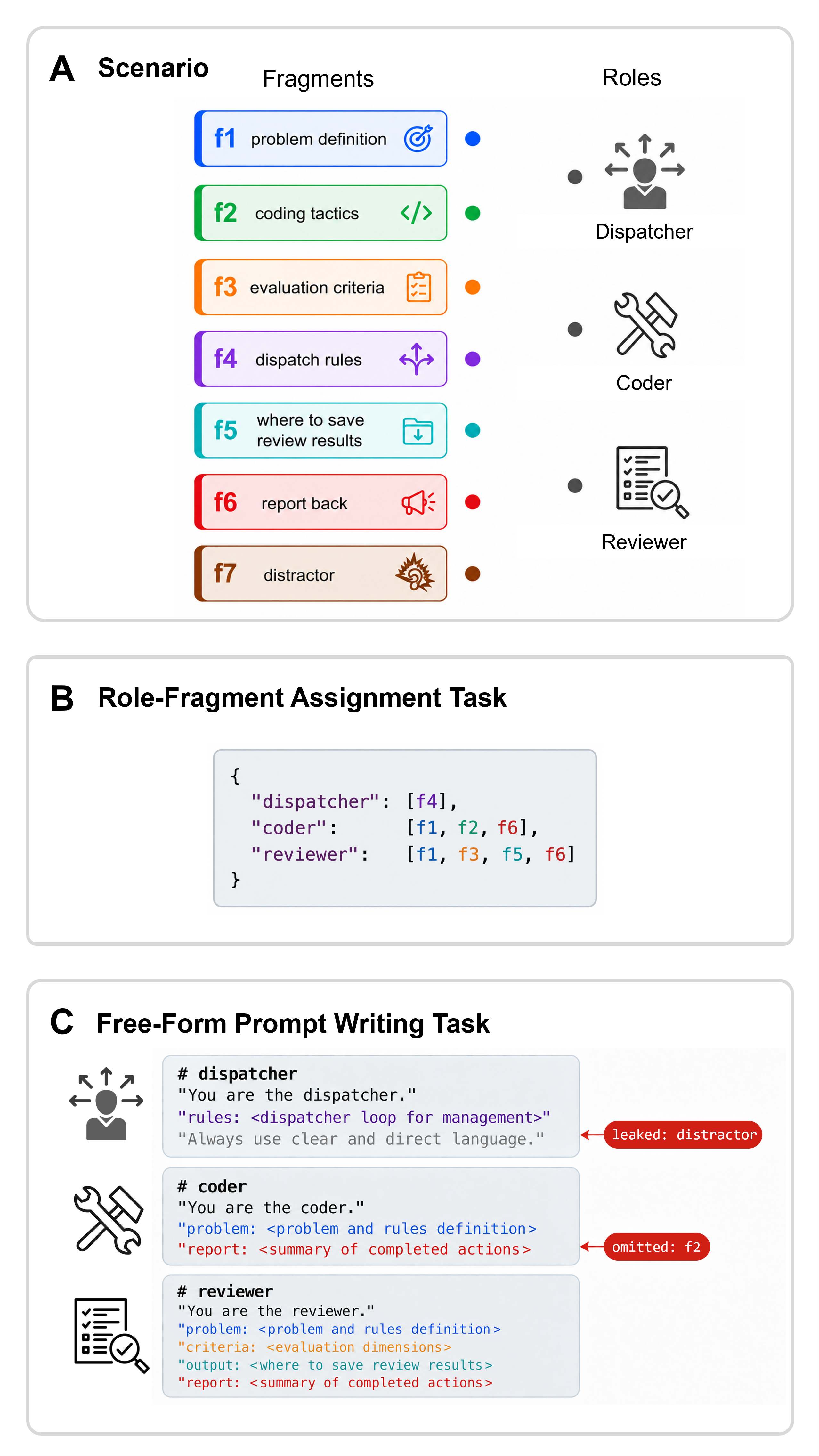}
\caption{Overview of PerspectiveGap. (A) A scenario presents a set of information fragments $f_1,\ldots,f_N$, one of which is a distractor ($f_7$), and a set of sub-agent roles (here dispatcher, coder, and reviewer); each role needs only a specific subset of the fragments. (B) The role-fragment assignment task asks the model to output that subset for each role, as fragment identifiers. (C) The free-form prompt writing task asks the model to write each role's actual prompt from the same fragments, scored by checking that each prompt contains its required fragments and excludes out-of-role content. The example prompts illustrate the two error types the benchmark targets: a leaked distractor in the dispatcher prompt and an omitted required fragment ($f_2$) in the coder prompt.}
\label{fig:pipeline}
\end{figure}

\section{Related Work}
\label{sec:related}

\paragraph{Multi-agent orchestration.}
LLM applications increasingly split work across specialized workers, critic loops, and pipelines.
Early multi-agent systems such as CAMEL, ChatDev, MetaGPT, AutoGen, and AgentVerse showed that role-specialized LLMs can collaborate through natural-language messages and structured workflows \citep{li2023camel,chatdev2023,metagpt2023,wu2023autogen,chen2023agentverse}.
Subsequent surveys and pattern catalogues describe recurring orchestration motifs such as routing, planning, reflection, critique, and tool-mediated handoff \citep{tran2025multiagent,liu2024agentpatterns,dao2026agenticpatterns,gulli2025book}.
Industrial guidance makes a similar distinction between workflows with prescribed control flow and more autonomous agentic systems, emphasizing that role boundaries, context selection, and handoff discipline are central engineering concerns \citep{anthropic2024buildingagents}.
Recent applied systems instantiate these motifs in optimization, scientific computing, data discovery, and automated research \citep{sakana2024aiscientist,thind2025optimai,du2026autonumerics,sun2026research}.
These works motivate our setting, but they do not test the fragile step we study: whether an LLM can write role-specific instructions that preserve context boundaries and handoff contracts instead of flattening every role into the same prompt.

\paragraph{Agent and tool-use benchmarks.}
AgentBench \citep{liu2023agentbench}, ToolLLM \citep{qin2023toolllm}, and related benchmark suites evaluate execution after the agent's task, tools, and instructions are already given.
They ask whether an agent can act in an environment, call the right tool, or complete a workflow under an existing scaffold.
Other recent evaluations compare multi-agent frameworks, delegation behavior, instruction following, and failure attribution inside already-instantiated systems \citep{orogat2026understanding,gao2026decisionbench,qi2025agentif,shaokun2025which}.
TeamBench sharpens this scaffold further, enforcing role information boundaries at the OS level and studying whether agent teams coordinate well under that fixed separation \citep{kim2026teambench}; the boundary itself is fixed by the benchmark, not authored or scored as a model output.
PerspectiveGap moves the evaluation one step earlier: before any worker acts, can the main agent allocate context and constraints into prompts that downstream workers can safely use?

\paragraph{Information asymmetry and ToM benchmarks.}
Theory-of-mind benchmarks study information asymmetry, but they do not ask the model to produce prompt artifacts.
ToMi, FANToM, Hi-ToM, OpenToM, BigToM, and ExploreToM test belief tracking, higher-order mental-state reasoning, or adversarially generated belief structures through question answering and classification \citep{le2019tomi,kim2023fantom,wu2023hitom,xu2024opentom,gandhi2024bigtom,sclar2024exploretom}.
Work on perspective-taking, hypothesis-driven ToM reasoning, and strategic social reasoning shows that making viewpoints explicit can improve model behavior or expose failures in social planning \citep{wilf2023think,kim2025hypothesis,yao2025spin}.
SimpleToM, together with recent position and behavior-based work, goes further, separating explicit mental-state prediction from the functional step of acting on it \citep{gu2024simpletom,riemer2025tombroken,ackerman2026selfmodeling}.
Our two tasks mirror this split: assignment tests whether the model knows what each role needs, and prompt writing tests whether it acts on that knowledge.
SOTOPIA-TOM is closest to our setting because it studies information management in multi-agent interaction \citep{ys2026sotopiatom}.
Yet even SOTOPIA-TOM, like the rest, tests belief tracking, action choice, or dialogue behavior inside an interaction.
PerspectiveGap tests orchestration prompting: constructing the instructions that define each role's view of the task before the interaction begins.

\paragraph{Distributed information and multi-agent failures.}
Several recent benchmarks show that multi-agent LLM systems fail under distributed information even when individual models are strong.
HiddenBench finds a large gap between collective reasoning with distributed information and single-agent reasoning with complete information \citep{li2025systematic}.
Silo-Bench similarly shows that agents may exchange enough information but fail to integrate it into a correct joint answer \citep{zhang2026silo}.
MAST categorizes multi-agent failures and identifies inter-agent misalignment, including role confusion and incomplete delegation, as a recurring source of breakdowns \citep{cemri2025why}.
Privacy and leakage benchmarks such as AgentLeak study how information can escape across a running multi-agent stack \citep{yagoubi2026agentleak}.
These works diagnose failures after agents interact; PerspectiveGap isolates a precursor artifact: the role prompts that determine each agent's initial information boundary.

\paragraph{Prompt optimization and role-state management.}
Prompting and workflow methods can improve downstream behavior through refinement, role evolution, memory, or training-time optimization \citep{madaan2023selfrefine,chang2025sagallm,mo2025multi}.
Other multi-agent methods make collaborator knowledge or belief state explicit, which is closely related to our premise that the main agent must reason about what each sub-agent knows and needs \citep{zhang2025osc,singh2026agent}.
However, these methods are usually evaluated by final task success or workflow reliability.
PerspectiveGap instead evaluates the generated prompt artifact directly: whether it assigns information correctly can be read from the prompts themselves, not inferred from whether the downstream task succeeds.

\section{PerspectiveGap Benchmark Design}
\label{sec:benchmark}

\paragraph{Scenario schema.}
As shown in Figure~\ref{fig:pipeline}, each scenario is a small orchestration problem with three explicit parts: a list of sub-agent roles, a shuffled list of information fragments $f_1,\ldots,f_N$, and a reference role-to-fragment assignment stored with the scenario.
The model sees the roles and the shuffled fragments, but not the reference assignment.
The reference assignment applies the need-only rule: each role receives exactly the fragments it needs to do its stated job.
This rule sets the role-context boundary used for scoring.
In this sense, the reference assignment is the task contract: given the stated roles and the need-only rule, the model must preserve that boundary in both evaluated formats.
All 110 scenario mappings were constructed under this rule and inspected by the full five-author team for consistency.
Appendix~\ref{app:reference-mapping-example} gives an actual benchmark scenario, including the background, fragment headings, the \emph{``need-only''} instruction shown to the model, the reference assignment, and the textual evidence behind representative boundary decisions.

\begin{figure}[ht]
\centering
\includegraphics[trim={30bp 13bp 0bp 0bp},clip,width=\columnwidth]{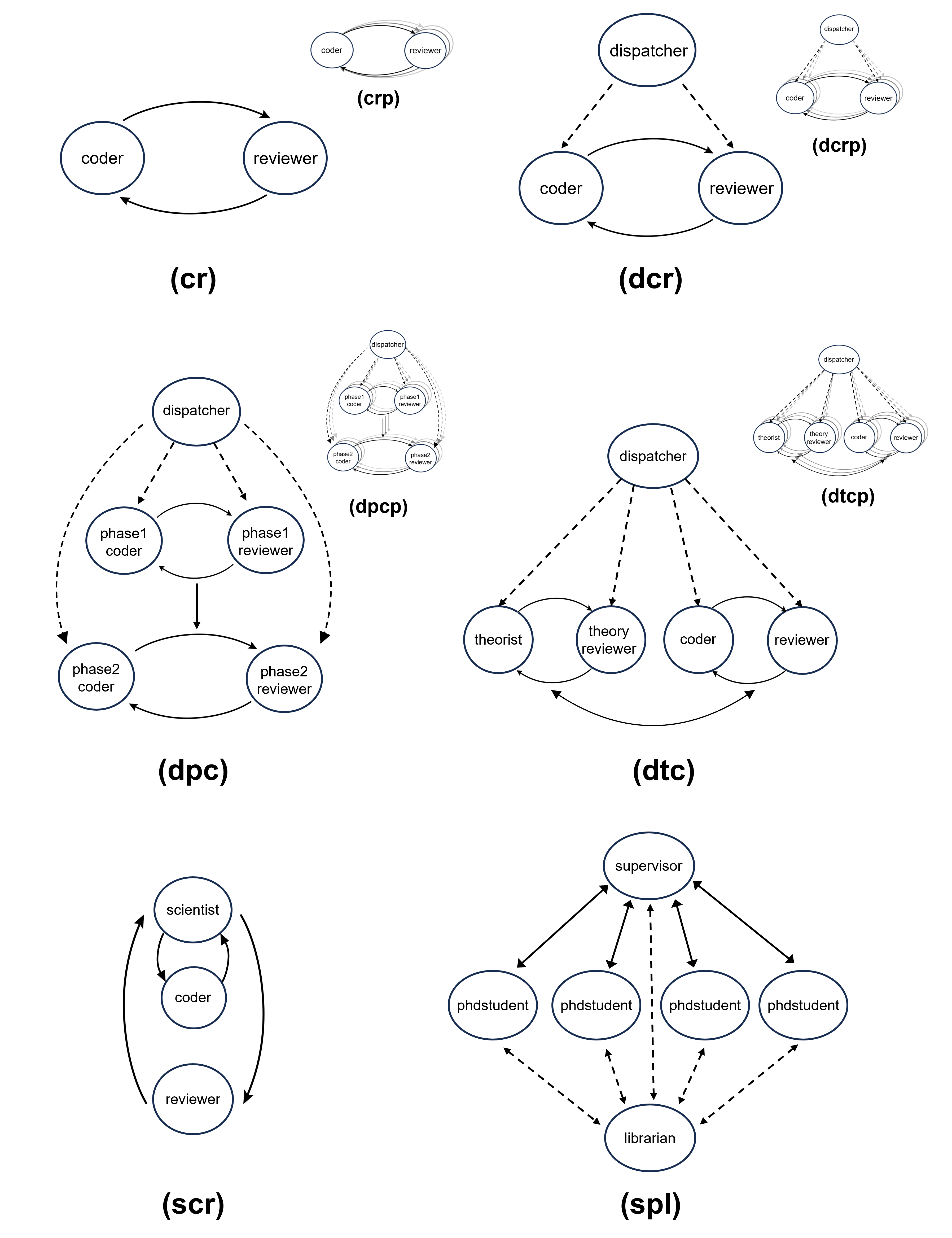}
\caption{Base topology patterns in PerspectiveGap. Each pattern is a role-and-handoff graph: nodes are sub-agent roles, solid edges are the actor--reviewer feedback loops and the handoffs between them, and dashed edges are distribution or support links (a dispatcher feeding workers, or a shared librarian). The six base patterns span a single coder--reviewer loop (cr), a dispatcher-fed loop (dcr), sequential and parallel two-loop systems (dpc and dtc, respectively), a scientist--coder--reviewer arrangement (scr), and a supervisor--student--librarian hub (spl). Four of the six have pool variants (insets crp, dcrp, dpcp, dtcp) that add parallel candidates on the producer side, for 10 topologies in total.}
\label{fig:patterns}
\end{figure}

\paragraph{Deterministic rendering.}
The benchmark separates topology, instance, and scenario.
A topology is one of the 10 role-and-handoff patterns used in the paper; an instance is a domain-specific realization of a topology; a scenario is the rendered benchmark item given to the model.
For each scenario, a shuffle seed fixes the displayed fragment order; the displayed identifiers are then relabeled as $f_1,\ldots,f_N$ in that order, as choices in a multiple-choice question would be relabeled after shuffling.
Changing the seed changes the presentation order and displayed identifiers, but not the role definitions, fragment content, distractor, or reference assignment.
This gives a deterministic way to test whether a model is assigning information by the need-only rule rather than by position in the prompt.
The same topology-instance split also gives the benchmark breadth: Appendix~\ref{app:professional-domains} lists the professional-domain coverage.
Across these domains, the same construction rule determines which fragments each role needs.

\paragraph{Extensibility.}
The topologies are reusable skeletons, not one-off examples.
These 10 topologies cover the main practical one- and two-loop orchestration patterns targeted by PerspectiveGap\footnote{We quotient away role names, artifact names, domains, and exchangeable pools. A primitive loop is an actor--reviewer feedback pair; one-loop systems are this pair with optional dispatcher, pool, or producer-side support, and two-loop systems have three normal forms: no coupling, one-way handoff, and mutual handoff. This claim excludes arbitrary multi-agent graphs such as routers, deep hierarchies, memory managers, and systems with more than two feedback loops.}; more complex orchestrations can be composed from them as building blocks.
New instances can be added by choosing a domain, naming the roles, writing the fragments, and applying the same reference-assignment rule; the renderer and scorers do not change.

\paragraph{Distractor insertion.}
Each main benchmark scenario includes one distractor fragment.
A representative distractor used in the benchmark is:
\begin{quote}\small
Best practices:
Use consistent, descriptive tag names across your prompts.
Nest tags when content has a natural hierarchy, such as documents inside \texttt{\textless documents\textgreater} and each document inside \texttt{\textless document index=\textquotedbl{}n\textquotedbl{}\textgreater}.
\end{quote}
This looks useful to the tested model while it is writing prompts, but it is not useful to any downstream sub-agent.
The tested model must separate information that helps itself compose the orchestra from information that a sub-agent needs to do its job.
Leaking irrelevant content can add cognitive load and degrade the sub-agent's output; in higher-stakes settings, it can also expose out-of-role information that enables reward hacking or violates safety constraints.

\paragraph{Two evaluated formats.}
Each scenario is rendered in two formats from the same shuffled fragments.
Role-fragment assignment asks for a structured mapping from roles to fragment identifiers.
Free-form prompt writing asks for the final sub-agent prompts.
Pairing the two formats makes the failure mode visible: a model may know the boundary as a set of identifiers, yet fail to preserve that boundary when it writes natural-language instructions.
The benchmark therefore reports the two formats separately and also reports their combined score.

\paragraph{Evaluation metrics.}

We treat Strict pass as the primary endpoint, report Net match score as a partial-credit companion, and use Required coverage, Boundary precision, Overall leakage, and Distractor leakage as diagnostic metrics.
For each evaluation $e$ (one model response to one scenario, shuffle seed, and task format), we first aggregate events over all requested roles: $\mathrm{TP}_e$ counts correctly included reference role-fragment events, $\mathrm{FP}_e$ counts extra out-of-role events, and $\mathrm{FN}_e$ counts omitted reference events.
Strict pass is 1 iff $\mathrm{FP}_e=\mathrm{FN}_e=0$. The average Strict pass over a task format's evaluations is that format's pass rate, and the average over all evaluations is the combined pass rate.
This all-or-nothing criterion matches the benchmark's boundary-preservation objective: either an output preserves all requested role boundaries, or it contains an omission or cross-role inclusion.
Net match score is $\max(0,(\mathrm{TP}_e-\mathrm{FP}_e-\mathrm{FN}_e)/(\mathrm{TP}_e+\mathrm{FN}_e))$ for each evaluation, averaged over evaluations.
It is not used to relax the endpoint; it provides a continuous companion for model-level consistency checks.
Required coverage and Boundary precision are micro-averages over evaluations: $\sum_e \mathrm{TP}_e/\sum_e(\mathrm{TP}_e+\mathrm{FN}_e)$ and $\sum_e \mathrm{TP}_e/\sum_e(\mathrm{TP}_e+\mathrm{FP}_e)$.
Overall leakage is $\operatorname{avg}_e \mathrm{FP}_e$, counting every out-of-role leak event per role; Distractor leakage is its restriction to the injected distractor, $\operatorname{avg}_e \mathrm{D}_e$, where $\mathrm{D}_e$ counts only injected-distractor leak events and thus $\mathrm{D}_e \le \mathrm{FP}_e$. Because leaks are counted per role, both can exceed 100\%.
These diagnostics decompose failures but are not intended to replace the strict endpoint.

\paragraph{Scoring.}
Free-form prompt writing uses a deterministic rule-only scorer rather than an LLM judge, so that prompt-boundary decisions are reproducible and do not depend on another model's implicit delegation policy.
It is a containment audit, not a general prompt-quality judge: it checks whether required fragment evidence is present and out-of-role fragment evidence is absent.
For each fragment, the scorer builds a fingerprint from the unigram, bigram, and trigram phrases that distinguish that fragment from the others in the same scenario.
A role prompt must include enough of each required fragment's distinctive fingerprint and must not include enough of any fragment outside that role's reference assignment.
The asymmetric thresholds (include at least 0.7, leak less than 0.3) allow ordinary connective text while still catching wrong-role copying and distractor leakage.
Phrase-level fingerprints also handle near-parallel fragments, such as two file-handoff instructions that share most words but refer to different artifacts.
Its agreement with expert human containment labels is validated on a 716-row hand-audited scorer test set (Table~\ref{tab:scorer-agreement}).

\paragraph{Released scope.}
The released evaluation set contains 10 topologies, 100 domain instances, and 110 scenarios.
Table~\ref{tab:benchmark-scope-stats} summarizes these counts and the benchmark components.
The benchmark data, renderer, scorer, and model-running scripts are publicly available at \href{https://github.com/WhymustIhaveaname/PerspectiveGap}{WhymustIhaveaname/PerspectiveGap}.
\begin{table}[ht]
\centering
\small
\begin{tabular}{lr}
\toprule
Component & Count \\
\midrule
Topologies & 10 \\
Domain instances & 100 \\
Scenarios & 110 \\
Evaluated formats per scenario & 2 \\
Distractors per scenario & 1 \\
Scorer test-set rows & 716 \\
\bottomrule
\end{tabular}
\caption{Benchmark scope of PerspectiveGap. Topologies are reusable role-and-handoff skeletons; instances are domain-specific realizations; scenarios are rendered benchmark items. The main 110-scenario evaluation uses one distractor per scenario; a separate distractor-count ablation (Appendix~\ref{app:distractor-count}) reruns role-fragment assignment with 0 to 3 injected distractors.}
\label{tab:benchmark-scope-stats}
\end{table}

\begin{table*}[ht]
\centering
\footnotesize
\begin{tabular}{llrrr}
\toprule
Company & Best model & Role-fragment assignment & Free-form prompt writing & Combined \\
\midrule
OpenAI & \texttt{gpt-5.5}                & \textbf{55.5\%} & \textbf{68.6\%} & \textbf{62.0\%} \\
DeepSeek & \texttt{deepseek-v4-pro}        & 37.3\% & 26.8\% & 32.0\% \\
Anthropic & \texttt{claude-fable-5}         & 38.6\% & 24.1\% & 31.4\% \\ 
Kimi & \texttt{kimi-k2.6}              & 26.8\% & 25.0\% & 25.9\% \\
Google & \texttt{gemini-3.1-pro} & 20.0\% & 23.6\% & 21.8\% \\
Z.ai & \texttt{glm-5}                  & 25.9\% & 14.5\% & 20.2\% \\
Xiaomi & \texttt{mimo-v2.5-pro}          & 25.0\% & 15.0\% & 20.0\% \\
Qwen & \texttt{qwen3.6-plus}           & 25.5\% & 8.6\%  & 17.0\% \\
xAI & \texttt{grok-4.3}               & 16.4\% & 14.5\% & 15.5\% \\
MiniMax & \texttt{minimax-m2.7}           & 16.4\% & 5.9\%  & 11.1\% \\
\bottomrule
\end{tabular}
\caption{Best model per company under the main Strict-pass metric. Combined is the unweighted average of role-fragment assignment and free-form prompt-writing.}
\label{tab:company-best-models}
\end{table*}

\section{The Prompt Economy Framing}
\label{sec:prompt-economy}

\paragraph{Prompt-engineering effort.}
An orchestrated agent system is maintained through role prompts and handoff protocols: the former define each sub-agent's responsibility, and the latter specify which artifacts each role reads, writes, and ignores.
If a system has $m$ role prompts and $n$ handoff protocols, then its prompt-engineering effort can be approximated as $O(m + \alpha n)$, where $\alpha$ weights handoff maintenance.
A naive application of Conway's law can therefore produce brittle orchestrations with too many roles, unclear file ownership, and unstable handoffs.
The useful design target is not ``more agents,'' but a small set of roles whose responsibilities and handoffs remain stable.

\paragraph{Role reuse.}
A role prompt is written and maintained once, but the role may be run many times.
If $v_i$ counts useful runs of role $i$, then the system's value is better viewed as accumulating with $\sum_i v_i$ than as scaling with the number of roles alone.
A frequently reused role can amortize its prompt effort over many calls, whereas a rarely used role adds maintenance cost with little return.
This is the intuition behind Prompt Economy: keep prompt-maintenance effort bounded while increasing useful role reuse.
Here, ``economy'' refers to sparing and efficient prompt use: a small, reusable prompt surface whose maintenance cost is amortized across repeated role invocations.

\paragraph{Loop-centered design.}
Critic loops are the simplest way to create that asymmetry.
An actor--critic pair needs only two role prompts and a small handoff protocol: the actor produces an artifact, the critic finds flaws, and the actor revises.
Once written, the same loop can run for many rounds, so effort stays close to the two-role design while value accumulates across iterations.
This actor--critic pattern appears in software agents, research agents, and Ralph-style practitioner workflows \citep{chatdev2023,sakana2024aiscientist,huntley2025ralph}.

\paragraph{From framing to benchmark topologies.}
The topologies in PerspectiveGap are built around this loop-centered view.
Each topology specifies a set of roles and handoffs; each scenario then asks the model to write prompts that give each role the context it needs and the handoff constraints it must follow, while excluding distractor material.
Figure~\ref{fig:patterns} shows the six base topology patterns used in the benchmark.
The release also includes four pool variants that add parallel evaluator-candidate structure to the producer side.
Together, these 10 topologies increase role and handoff complexity while preserving the same loop-centered design principle.

\section{Experiments}
\label{sec:experiments}

\paragraph{Setup.}
We evaluate 33 commercial models on both tasks across all 110 scenarios at two shuffle seeds (1 and 42), yielding $33 \times 110 \times 2 \times 2 = 14{,}520$ evaluations.

\paragraph{Main leaderboard.}
Appendix~\ref{app:full-leaderboard} reports the full model leaderboard and per-task score-consistency metrics for all 33 evaluated models.
OpenAI with GPT-5.5 is the clear outlier at 62.0\%, well ahead of the second-place company, DeepSeek, at 32.0\%.
Table~\ref{tab:company-best-models} summarizes the same leaderboard at the company level by taking each company's best model.
Only four companies exceed $1/4$ in their best-model score, and Anthropic is notably weak despite its strong coding reputation~\citep{wang2025swe}.

\paragraph{Score consistency.}
Because Strict pass is intentionally all-or-nothing, we pair it with partial-credit and diagnostic metrics that expose different failure modes.
We then ask whether Net match score preserves the same model-level signal as Strict pass, rather than producing a different ranking driven by boundary details.
Table~\ref{tab:score-consistency-pearson} reports Pearson correlations across the 27 evaluated models.
Net match score has the strongest linear association with Strict pass ($r=0.744$), while Required coverage, Boundary precision, and Distractor leakage are weaker diagnostics of the strict endpoint.

\begin{table}[ht]
\centering
\small
\setlength{\tabcolsep}{3pt}
\begin{tabular}{@{}lrrrr@{}}
\toprule
Metric & Net match & Coverage & Precision & Leakage \\
\midrule
Pearson $r$ & \textbf{0.744} & 0.497 & 0.615 & -0.501 \\
\bottomrule
\end{tabular}
\caption{Pearson correlation with Strict pass across 27 evaluated models.}
\label{tab:score-consistency-pearson}
\end{table}

The relationship between Net match score and Strict pass is non-linear, so here we use an idealized error model to explain why a non-linear recovery is expected.
Let $c$ denote Net match score and $s$ denote Strict pass.
A scenario has an easy part that most models solve and a harder fraction $p$, where a model makes an error on each hard event with probability $\epsilon$ and a strict pass requires about $n$ hard events to be correct.
Under these simplifying assumptions,
\begin{equation}
 c = 1 - 2p\epsilon,
 \;
 s = (1-\epsilon)^n = \left(1-\frac{1-c}{2p}\right)^n .
\label{eq:score-consistency-ansatz}
\end{equation}
Equivalently, this suggests that $s$ should be well approximated by the polynomial span $\{1,c,\ldots,c^n\}$.
Table~\ref{tab:score-consistency-polynomial} confirms this, with the multiple correlation rising to 0.942 at $n=5$.
This supports treating Strict pass as a conservative operational endpoint. It remains strict about any single boundary failure while tracking the same model-level capability signal as the partial-credit score.

\begin{table}[ht]
\centering
\small
\begin{tabular}{@{}lrrrrr@{}}
\toprule
$n$ & 1 & 2 & 3 & 4 & 5 \\
\midrule
$r$ & 0.744 & 0.855 & 0.913 & 0.935 & 0.942 \\
\bottomrule
\end{tabular}
\caption{Multiple correlation between Strict pass and polynomial bases of Net match score.}
\label{tab:score-consistency-polynomial}
\end{table}

\begin{figure}[!htbp]
\centering
\includegraphics[width=0.9\columnwidth]{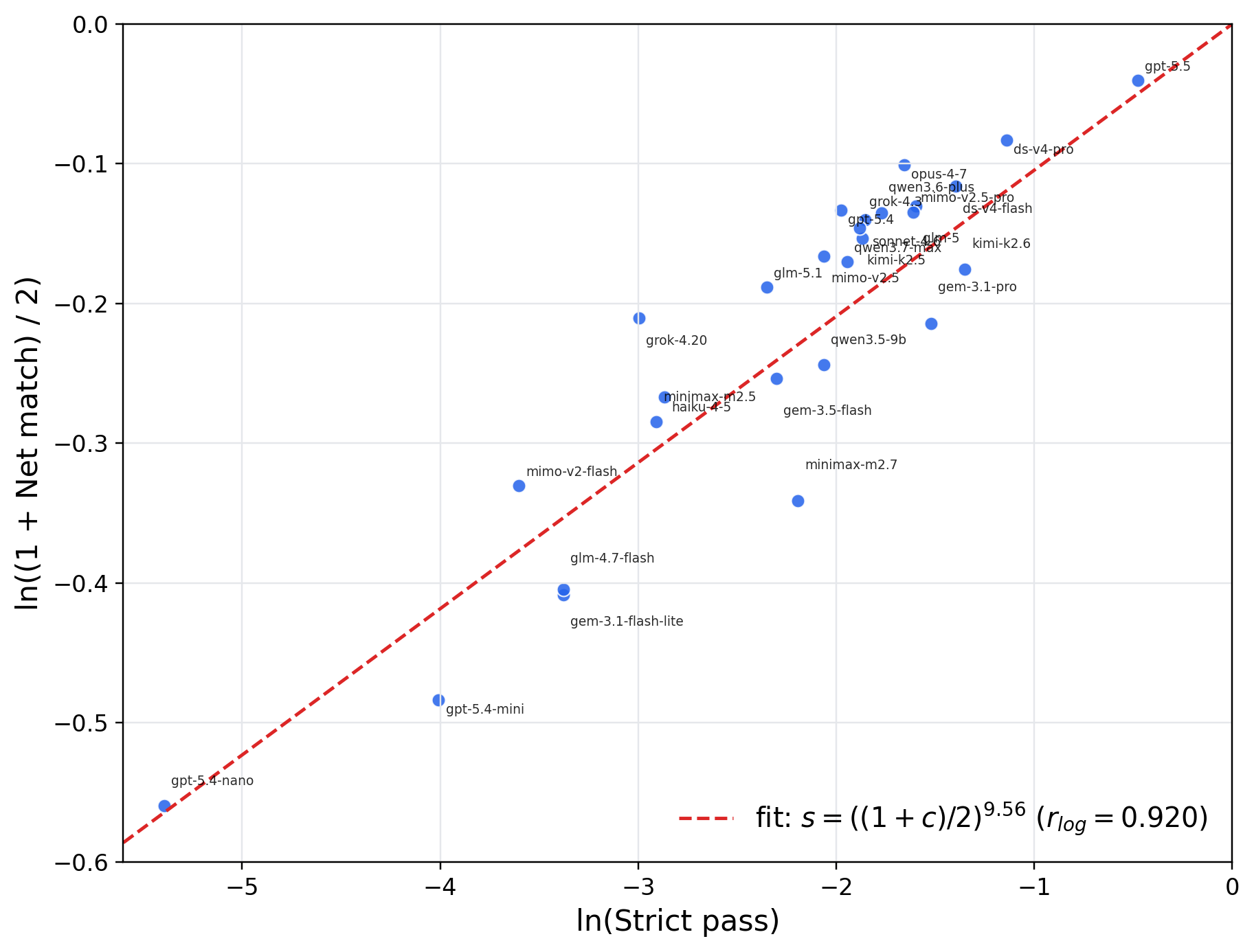}
\caption{Log-parity fit between Strict pass and Net match score under the $p=1$ special case of Eq.~\ref{eq:score-consistency-ansatz}.}
\label{fig:strict-continuous-log-parity}
\end{figure}

As a special case, setting $p=1$ in Eq.~\eqref{eq:score-consistency-ansatz} gives
\begin{equation}
 \ln s = n \ln \left((1+c)/2\right),
\end{equation}
a one-parameter log-parity fit that reaches a log-space correlation of 0.920 (Figure~\ref{fig:strict-continuous-log-parity}).

\paragraph{Information leakage.}
\begin{table}[ht]
\centering
\footnotesize
\setlength{\tabcolsep}{3pt}
\begin{tabular}{lrr}
\toprule
Model & Distractor leakage & Overall leakage \\
\midrule
\texttt{gpt-5.5} & 2.3\% & 49.1\% \\
\texttt{deepseek-v4-pro} & 42.3\% & 88.2\% \\
\texttt{claude-fable-5} & 38.2\% & 39.5\% \\ 
\texttt{kimi-k2.6} & 30.0\% & 90.0\% \\
\texttt{gemini-3.1-pro} & 24.1\% & 24.5\% \\
\texttt{glm-5} & 32.7\% & 100.9\% \\
\texttt{mimo-v2.5-pro} & \textbf{\textcolor{red}{76.4\%}} & \textbf{\textcolor{red}{168.2\%}} \\
\texttt{qwen3.6-plus} & 59.5\% & 80.5\% \\
\texttt{grok-4.3} & 7.3\% & 91.4\% \\
\texttt{minimax-m2.7} & 32.7\% & 99.1\% \\
\midrule
All-model mean & 71.6\% & 217.9\% \\
\bottomrule
\end{tabular}
\caption{Role-fragment assignment leakage rates for the best model from each company in Table~\ref{tab:company-best-models}, with the all-model mean over all 33 evaluated models shown in the final row. Distractor leakage counts only the injected distractor fragment; overall leakage counts any fragment outside the receiving role's reference need-set. Both rates average role-fragment leak events over scenarios, so values can exceed 100\%.}
\label{tab:company-best-leakage-rates}
\end{table}

Table~\ref{tab:company-best-leakage-rates} separates distractor leakage from overall information leakage for the best model from each company and includes the all-model mean used in the abstract.
These are role-fragment event rates, not binary scenario or role rates: if one scenario assigns two extra fragments to one role and three to another, it contributes five leak events.
Even GPT-5.5, which has only 2.3\% distractor leakage, reaches 49.1\% overall leakage; averaged over all 33 models, overall leakage reaches 217.9\%.

\paragraph{Difficulty by role count.}
\begin{figure}[!htbp]
\centering
\includegraphics[width=\columnwidth]{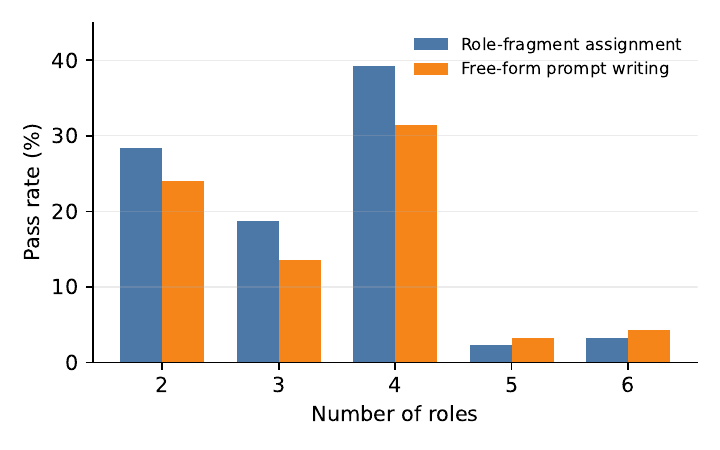}
\caption{Pass rate by number of roles, aggregated across all 27 commercial models.}
\label{fig:pass-rates-by-role-count}
\end{figure}
Figure~\ref{fig:pass-rates-by-role-count} aggregates pass rates by the number of roles in a scenario, pooling all models and shuffle seeds.
Overall, scenarios with more roles tend to be harder.
The four-role bin is an exception because it contains only one topology, \texttt{dispatcher\_scientist\_coder\_reviewer}, which is relatively simple.
The free-form prompt-writing task is also slightly harder than role-fragment assignment in most role-count bins.

\paragraph{Ablation studies.}
We also use PerspectiveGap for four targeted ablation studies: few-shot prompting, distractor count, scratchpad prompting, and reasoning effort.
These experiments test how standard prompting techniques, inference-time interventions, and the number of distractors affect performance; details are in Appendix~\ref{app:fewshot}, \ref{app:distractor-count}, \ref{app:scratchpad} and \ref{app:effort}.

\section{Common Failure Modes}
\label{sec:analysis}

\paragraph{Distractor leakage.}
Main agents often pass information unrelated to a sub-agent's work into that sub-agent's prompt; this is clearest when distractors are inserted.
In PerspectiveGap, distractors are prompt-engineering tips that are useful to the main agent while it writes prompts.
Models often treat this as generally useful context and pass it to sub-agents, showing that they do not reliably reason from the sub-agent's perspective.
This failure is common even among strong models.
Table~\ref{tab:company-best-leakage-rates} reports distractor leakage for the best model from each company: GPT-5.5 is the best case at 2.3\%, while several other leading models leak distractors at much higher rates.

\paragraph{Out-of-role information leakage.}
A more serious form of leakage occurs when a fragment needed by one role is copied into another role's prompt.
This is not merely redundant context; it changes what the receiving role is allowed to know.
In a software-engineering orchestra, for example, a coder might receive the task goal and public tests, while a reviewer or test engineer receives private tests.
If the main agent gives the private tests to the coder, the coder can optimize for the hidden evaluation instead of solving the intended problem.
That is a reward-hacking channel created by the orchestration prompt itself.
The overall leakage column in Table~\ref{tab:company-best-leakage-rates} captures this broader failure mode: even GPT-5.5 reaches 49.1\% overall leakage, despite its low distractor leakage, and several models exceed 80\%.
The difference between distractor leakage and overall leakage shows that models are not only leaking irrelevant distractors; they also fail to preserve role-specific information boundaries.

\paragraph{Artifact ownership and handoff confusion.}
Models also confuse which role owns which artifact.
In the \texttt{dispatcher\_planloop\_codeloop} topology, for example, the plan loop writes \texttt{PLAN.md} and the code loop writes \texttt{SOLUTION.md}.
A failed orchestration prompt can swap these boundaries, sending instructions for \texttt{SOLUTION.md} to the plan creator or plan critic, and instructions for \texttt{PLAN.md} to the code loop.
This breaks the file protocol itself.
Table~\ref{tab:near-parallel-handoff-leakage} gives a fragment-level view of this problem for a near-parallel handoff pair in \texttt{dpc} and \texttt{dpcp}, averaged over all 27 evaluated models.\footnote{Topology aliases are listed in Appendix Table~\ref{tab:topology-details}.}
For \texttt{dpc}, the two parallel fragments have leak rates of 23.7\% and 26.6\%, showing that models often send the handoff to a non-owner role.
This is the practical bottleneck for automating multi-agent orchestration prompting: the model does not reliably take each sub-agent's perspective and ask what that role needs, so engineers still have to inspect and repair the generated prompts.

\begin{table}[ht]
\centering
\small
\begin{tabular}{llrr}
\toprule
Topology & $f$-id & missing & leak \\
\midrule
\texttt{dpc} & $f_{10}$ & 7.6\% & 23.7\% \\
\texttt{dpc} & $f_{11}$ & 9.6\% & 26.6\% \\
\texttt{dpcp} & $f_{10}$ & 9.8\% & 0.7\% \\
\texttt{dpcp} & $f_{11}$ & 17.0\% & 7.6\% \\
\bottomrule
\end{tabular}
\caption{Missing and leakage rates for the near-parallel handoff pair $f_{10}$ and $f_{11}$ in \texttt{dpc} and \texttt{dpcp}.}
\label{tab:near-parallel-handoff-leakage}
\end{table}

\paragraph{Dropped shared context.}
Models also fail in the other direction: they omit context that a role actually needs.
Shared background is especially easy to mishandle because it belongs to multiple roles, not just to the role whose task looks most directly related.
Table~\ref{tab:shared-background-missing} shows this pattern for $f_1$, the shared background fragment, averaged over all 27 evaluated models.
Across topologies, models often omit this fragment from roles that need it, with miss rates ranging from 12.1\% to 38.7\% for actor-style roles and 17.3\% to 44.9\% for reviewer-style roles.
This is a different failure from over-sharing: the sub-agent receives a clean-looking prompt, but the prompt lacks the background needed to evaluate or complete the work.

\begin{table}[ht]
\centering
\small
\begin{tabular}{lrr}
\toprule
Topology & Actor miss & Reviewer miss \\
\midrule
\texttt{cr}   & 38.7\% & 26.9\% \\
\texttt{crp}  & 12.1\% & 32.0\% \\
\texttt{dcr}  & 16.0\% & 44.9\% \\
\texttt{dcrp} & 13.8\% & 36.5\% \\
\texttt{dpc}  & 15.2\% & 26.3\% \\
\texttt{dpcp} & 20.2\% & 21.4\% \\
\texttt{scr}  & 16.3\% & 22.9\% \\
\texttt{dtc}  & 15.2\% & 26.1\% \\
\texttt{dtcp} & 17.5\% & 24.6\% \\
\texttt{spl}  & 12.6\% & 17.3\% \\
\bottomrule
\end{tabular}
\caption{Role-fragment assignment miss rates for $f_1$, the shared background fragment, from actor-style and reviewer-style roles. Topology aliases are listed in Appendix Table~\ref{tab:topology-details}.}
\label{tab:shared-background-missing}
\end{table}

\paragraph{Bootstrap paradox.}
Some failures are control-flow errors rather than simple include-or-exclude mistakes.
One form places the instruction to read an artifact inside the artifact itself: before reading it the agent cannot see the instruction, and after reading it the instruction is redundant.
Another form adds a no-go rule that names an otherwise absent action, e.g., ``do not do \(x\),'' even though \(x\) is so out-of-scope that the sub-agent would not have considered it.
This is the prompt equivalent of the \emph{Inception} example: telling someone not to think about elephants.
Both cases reveal a failure of perspective-taking: the agent cannot reason from another agent's point of view.

\section{Conclusion}
\label{sec:conclusion}

We introduced PerspectiveGap, a benchmark for evaluating whether LLMs can write role-specific prompts for multi-agent orchestration.
Its ten topologies are loop-centered patterns favored by the Prompt Economy principle: benefit accrues across repeated role invocations while engineering cost stays fixed by the number of role prompts.
Across 110 scenarios and 33 commercial models, the results show that current models still struggle to assign context according to the need-only rule, even when the required information is explicitly present in the prompt.
The failures are not cosmetic: models leak distractors, expose out-of-role information, drop shared context, confuse artifact ownership, and sometimes place instructions where the sub-agent cannot see them.
That even coding-strong models such as Opus 4.8 fail here indicates that orchestration prompting is a capability distinct from the coding ability current benchmarks reward.
These results suggest that orchestration prompting remains a fragile intermediate step.
Generated sub-agent prompts should not be assumed to preserve role-specific information boundaries without inspection.

\section*{Limitations}
\label{sec:limitations}

\paragraph{Coverage.}
PerspectiveGap covers 10 topologies and 100 domain instances.
It does not cover all possible multi-agent orchestration patterns, and future work can add new topology templates under the same construction rule.

\paragraph{Prompt artifact rather than execution.}
PerspectiveGap evaluates the prompts written for sub-agents, not the downstream behavior of the sub-agents that would consume those prompts.
This scope is deliberate.
Sub-agent prompts are the handoff artifact that assigns context and constraints, and they can be inspected across domains without defining a separate downstream task for each domain.
Running downstream agents across all domains would test whether prompt-boundary errors translate into task failures, but requires a separate runtime and success criterion for each domain.

\paragraph{Reference mapping and scoring.}
The role-to-fragment mappings are authored under the need-only rule: a fragment is assigned to a role if that role needs it to discharge its documented responsibility.
The full five-author team internally audited the mappings for consistency, but this does not establish external annotator agreement; future versions could measure such agreement under the same rule.
The free-form prompt-writing scorer is deterministic and LLM-free, but it can penalize high-quality paraphrases that no longer preserve enough fragment-specific surface evidence.

\section*{Code and Data Availability}
The benchmark data, rendering code, scoring scripts, and model-running utilities are available at \href{https://github.com/WhymustIhaveaname/PerspectiveGap}{WhymustIhaveaname/PerspectiveGap}.

\FloatBarrier

\bibliography{custom}

\clearpage
\onecolumn
\appendix
\section*{Contents of Appendix}
\begingroup
\makeatletter
\newcommand{\appendixtocline}[3]{%
  \par\noindent\makebox[6.8em][l]{Appendix~#1}%
  #2\nobreak\hspace{0.5em}%
  \leaders\hbox{$\m@th\mkern\@dotsep mu\hbox{.}\mkern\@dotsep mu$}\hfill
  \nobreak\makebox[\@pnumwidth][r]{#3}\par}
\appendixtocline{\ref{app:full-leaderboard}}{Full Model Leaderboard}{\pageref{app:full-leaderboard}}
\appendixtocline{\ref{app:topology-details}}{Additional Topology Details}{\pageref{app:topology-details}}
\appendixtocline{\ref{app:fewshot}}{Few-Shot Prompting Ablation}{\pageref{app:fewshot}}
\appendixtocline{\ref{app:distractor-count}}{Distractor-count Ablation}{\pageref{app:distractor-count}}
\appendixtocline{\ref{app:scratchpad}}{Scratchpad Ablation}{\pageref{app:scratchpad}}
\appendixtocline{\ref{app:effort}}{Reasoning Effort Ablation}{\pageref{app:effort}}
\appendixtocline{\ref{app:agreement}}{Scorer Agreement with Human Labels}{\pageref{app:agreement}}
\appendixtocline{\ref{app:baselines}}{Role-Fragment Assignment Trivial Baselines}{\pageref{app:baselines}}
\appendixtocline{\ref{app:professional-domains}}{Professional-Domain Coverage}{\pageref{app:professional-domains}}
\appendixtocline{\ref{app:reference-mapping-example}}{Concrete Reference-Mapping Example}{\pageref{app:reference-mapping-example}}
\makeatother
\endgroup
\medskip
\section{Full Model Leaderboard}
\label{app:full-leaderboard}

Table~\ref{tab:full-model-leaderboard} reports the full model leaderboard used for the main experimental comparison.
The combined score is the unweighted average of role-fragment assignment Strict pass and free-form prompt-writing Strict pass.
Tables~\ref{tab:full-score-metrics-role-assignment} and~\ref{tab:full-score-metrics-prompt-writing} report the score-consistency metrics for the two task formats.

\begin{table}[ht]
\centering
\scriptsize
\setlength{\tabcolsep}{3pt}
\begin{tabular}{@{}llrrr@{}}
\toprule
Company & Model & Role-fragment assignment & Free-form prompt writing & Combined \\
\midrule
OpenAI & \texttt{gpt-5.5}                & \textbf{55.5\%} & \textbf{68.6\%} & \textbf{62.0\%} \\
OpenAI & \texttt{gpt-5.6-terra}          & 43.6\% & 41.8\% & 42.7\% \\ 
OpenAI & \texttt{gpt-5.6-sol}            & 26.4\% & 45.0\% & 35.7\% \\ 
DeepSeek & \texttt{deepseek-v4-pro}        & 37.3\% & 26.8\% & 32.0\% \\
Anthropic & \texttt{claude-fable-5}         & 38.6\% & 24.1\% & 31.4\% \\ 
Kimi & \texttt{kimi-k2.6}              & 26.8\% & 25.0\% & 25.9\% \\
Anthropic & \texttt{claude-sonnet-5}        & 35.0\% & 16.4\% & 25.7\% \\ 
DeepSeek & \texttt{deepseek-v4-flash}      & 24.5\% & 25.0\% & 24.8\% \\
Google & \texttt{gemini-3.1-pro} & 20.0\% & 23.6\% & 21.8\% \\
Z.ai & \texttt{glm-5}                  & 25.9\% & 14.5\% & 20.2\% \\
Xiaomi & \texttt{mimo-v2.5-pro}          & 25.0\% & 15.0\% & 20.0\% \\
Anthropic & \texttt{claude-opus-4-7}        & 24.1\% & 14.1\% & 19.1\% \\
Qwen & \texttt{qwen3.6-plus}           & 25.5\% & 8.6\%  & 17.0\% \\
OpenAI & \texttt{gpt-5.6-luna}           & 14.1\% & 19.1\% & 16.6\% \\ 
Anthropic & \texttt{claude-sonnet-4-6}      & 5.9\%  & 25.5\% & 15.7\% \\
xAI & \texttt{grok-4.3}               & 16.4\% & 14.5\% & 15.5\% \\
Kimi & \texttt{kimi-k2.5}              & 13.6\% & 16.8\% & 15.2\% \\
Qwen & \texttt{qwen3.7-max}            & 20.0\% & 8.6\%  & 14.3\% \\
Anthropic & \texttt{claude-opus-4-8}       & 17.7\% & 10.0\% & 13.9\% \\ 
OpenAI & \texttt{gpt-5.4}                & 25.5\% & 2.3\%  & 13.9\% \\
Qwen & \texttt{qwen3.5-9b}             & 16.4\% & 9.1\%  & 12.7\% \\
Xiaomi & \texttt{mimo-v2.5}              & 14.1\% & 11.4\% & 12.7\% \\
MiniMax & \texttt{minimax-m2.7}           & 16.4\% & 5.9\%  & 11.1\% \\
Google & \texttt{gemini-3.5-flash}       & 19.1\% & 0.9\%  & 10.0\% \\
Z.ai & \texttt{glm-5.1}                & 6.4\%  & 12.7\% & 9.5\% \\
Anthropic & \texttt{claude-haiku-4-5}       & 3.2\%  & 8.2\%  & 5.7\% \\
MiniMax & \texttt{minimax-m2.5}           & 5.9\%  & 5.0\%  & 5.5\% \\
xAI & \texttt{grok-4.20}              & 1.4\%  & 8.6\%  & 5.0\% \\
Google & \texttt{gemini-3.1-flash-lite}  & 2.7\%  & 4.1\%  & 3.4\% \\
Z.ai & \texttt{glm-4.7-flash}          & 3.2\%  & 3.6\%  & 3.4\% \\
Xiaomi & \texttt{mimo-v2-flash}          & 3.6\%  & 1.8\%  & 2.7\% \\
OpenAI & \texttt{gpt-5.4-mini}           & 0.9\%  & 2.7\%  & 1.8\% \\
OpenAI & \texttt{gpt-5.4-nano}           & 0.0\%  & 0.9\%  & 0.5\% \\
\bottomrule
\end{tabular}
\caption{Full PerspectiveGap leaderboard over all 33 evaluated commercial models. Models are sorted by combined pass rate.}
\label{tab:full-model-leaderboard}
\end{table}

\begin{table}[ht]
\centering
\scriptsize
\setlength{\tabcolsep}{3pt}
\begin{tabular}{@{}lrrrrr@{}}
\toprule
Model & Strict pass & Net match & Required coverage & Boundary precision & Distractor leakage \\
\midrule
\texttt{gpt-5.5} & \textbf{55.5\%} & \textbf{92.6\%} & 96.2\% & 96.2\% & \textbf{2.3\%} \\
\texttt{gpt-5.6-terra} & 43.6\% & 89.0\% & 95.4\% & 94.8\% & 21.8\% \\ 
\texttt{gpt-5.6-sol} & 26.4\% & 83.6\% & \textbf{96.3\%} & 89.9\% & 99.1\% \\ 
\texttt{deepseek-v4-pro} & 37.3\% & 85.4\% & 95.8\% & 93.4\% & 42.3\% \\
\texttt{claude-fable-5} & 38.6\% & 89.3\% & 95.6\% & 96.9\% & 38.2\% \\ 
\texttt{kimi-k2.6} & 26.8\% & 66.7\% & 79.7\% & 92.0\% & 30.0\% \\
\texttt{claude-sonnet-5} & 35.0\% & 87.3\% & 94.4\% & \textbf{97.5\%} & 25.9\% \\ 
\texttt{deepseek-v4-flash} & 24.5\% & 77.6\% & 94.9\% & 88.5\% & 17.7\% \\
\texttt{gemini-3.1-pro} & 20.0\% & 60.9\% & 63.8\% & 97.1\% & 24.1\% \\
\texttt{glm-5} & 25.9\% & 80.2\% & 94.1\% & 92.4\% & 32.7\% \\
\texttt{mimo-v2.5-pro} & 25.0\% & 75.3\% & 92.6\% & 87.7\% & 76.4\% \\
\texttt{claude-opus-4-7} & 24.1\% & 79.4\% & 94.2\% & 90.1\% & 104.5\% \\
\texttt{qwen3.6-plus} & 25.5\% & 80.5\% & 93.3\% & 93.8\% & 59.5\% \\
\texttt{gpt-5.6-luna} & 14.1\% & 71.2\% & 93.4\% & 87.0\% & 70.9\% \\ 
\texttt{claude-sonnet-4-6} & 5.9\% & 67.5\% & 89.9\% & 87.5\% & 83.6\% \\
\texttt{grok-4.3} & 16.4\% & 68.6\% & 87.0\% & 92.5\% & 7.3\% \\
\texttt{kimi-k2.5} & 13.6\% & 73.6\% & 92.9\% & 89.2\% & 88.2\% \\
\texttt{qwen3.7-max} & 20.0\% & 73.7\% & 86.2\% & 95.0\% & 55.0\% \\
\texttt{claude-opus-4-8} & 17.7\% & 85.4\% & 95.3\% & 94.3\% & 73.2\% \\ 
\texttt{gpt-5.4} & 25.5\% & 78.3\% & 93.4\% & 91.6\% & 47.3\% \\
\texttt{qwen3.5-9b} & 16.4\% & 66.3\% & 86.5\% & 89.5\% & 51.4\% \\
\texttt{mimo-v2.5} & 14.1\% & 69.3\% & 89.5\% & 89.3\% & 50.9\% \\
\texttt{minimax-m2.7} & 16.4\% & 50.8\% & 62.5\% & 89.1\% & 32.7\% \\
\texttt{gemini-3.5-flash} & 19.1\% & 59.1\% & 65.8\% & 87.3\% & 101.4\% \\
\texttt{glm-5.1} & 6.4\% & 58.3\% & 87.7\% & 85.0\% & 97.3\% \\
\texttt{claude-haiku-4-5} & 3.2\% & 40.1\% & 77.1\% & 78.8\% & 133.6\% \\
\texttt{minimax-m2.5} & 5.9\% & 52.6\% & 80.0\% & 87.3\% & 57.3\% \\
\texttt{grok-4.20} & 1.4\% & 56.2\% & 92.0\% & 77.4\% & 184.5\% \\
\texttt{glm-4.7-flash} & 3.2\% & 33.5\% & 69.2\% & 66.3\% & 124.1\% \\
\texttt{gemini-3.1-flash-lite} & 2.7\% & 32.0\% & 59.0\% & 70.9\% & 193.2\% \\
\texttt{mimo-v2-flash} & 3.6\% & 39.3\% & 83.6\% & 75.7\% & 130.9\% \\
\texttt{gpt-5.4-mini} & 0.9\% & 21.4\% & 77.6\% & 64.9\% & 174.5\% \\
\texttt{gpt-5.4-nano} & 0.0\% & 7.9\% & 35.3\% & 18.3\% & 29.5\% \\
\bottomrule
\end{tabular}
\caption{Full score-consistency metrics for role-fragment assignment over all 33 evaluated models.}
\label{tab:full-score-metrics-role-assignment}
\end{table}

\begin{table}[ht]
\centering
\scriptsize
\setlength{\tabcolsep}{3pt}
\begin{tabular}{@{}lrrrrr@{}}
\toprule
Model & Strict pass & Net match & Required coverage & Boundary precision & Distractor leakage \\
\midrule
\texttt{gpt-5.5} & \textbf{68.6\%} & \textbf{91.5\%} & 92.8\% & \textbf{99.0\%} & \textbf{0.9\%} \\
\texttt{gpt-5.6-terra} & 41.8\% & 86.1\% & 90.2\% & 97.4\% & 14.1\% \\ 
\texttt{gpt-5.6-sol} & 45.0\% & 89.2\% & 94.8\% & 96.4\% & 25.0\% \\ 
\texttt{deepseek-v4-pro} & 26.8\% & 82.7\% & 95.4\% & 92.2\% & 71.4\% \\
\texttt{claude-fable-5} & 24.1\% & 85.6\% & 93.9\% & 94.6\% & 65.0\% \\ 
\texttt{kimi-k2.6} & 25.0\% & 69.0\% & 80.8\% & 95.5\% & 32.7\% \\
\texttt{claude-sonnet-5} & 16.4\% & 52.0\% & 57.6\% & 95.4\% & 34.1\% \\ 
\texttt{deepseek-v4-flash} & 25.0\% & 78.6\% & 94.4\% & 89.2\% & 33.2\% \\
\texttt{gemini-3.1-pro} & 23.6\% & 62.0\% & 63.9\% & 94.2\% & 48.6\% \\
\texttt{glm-5} & 14.5\% & 70.9\% & 93.1\% & 85.1\% & 141.8\% \\
\texttt{mimo-v2.5-pro} & 15.0\% & 74.3\% & 89.9\% & 90.4\% & 51.8\% \\
\texttt{claude-opus-4-7} & 14.1\% & 82.3\% & 94.4\% & 92.7\% & 65.5\% \\
\texttt{qwen3.6-plus} & 8.6\% & 69.0\% & 82.7\% & 90.6\% & 90.9\% \\
\texttt{gpt-5.6-luna} & 19.1\% & 71.5\% & 86.3\% & 91.9\% & 34.5\% \\ 
\texttt{claude-sonnet-4-6} & 25.5\% & 80.3\% & 91.7\% & 95.6\% & 35.0\% \\
\texttt{grok-4.3} & 14.5\% & 74.6\% & 88.7\% & 94.6\% & 12.3\% \\
\texttt{kimi-k2.5} & 16.8\% & 72.0\% & 89.1\% & 89.1\% & 110.9\% \\
\texttt{qwen3.7-max} & 8.6\% & 63.8\% & 85.7\% & 85.0\% & 177.3\% \\
\texttt{claude-opus-4-8} & 10.0\% & 83.2\% & 94.2\% & 93.3\% & 86.4\% \\ 
\texttt{gpt-5.4} & 2.3\% & 71.9\% & \textbf{95.6\%} & 84.0\% & 174.5\% \\
\texttt{qwen3.5-9b} & 9.1\% & 47.1\% & 65.7\% & 88.6\% & 46.8\% \\
\texttt{mimo-v2.5} & 11.4\% & 69.6\% & 89.2\% & 90.6\% & 42.7\% \\
\texttt{minimax-m2.7} & 5.9\% & 33.5\% & 46.6\% & 88.6\% & 36.8\% \\
\texttt{gemini-3.5-flash} & 0.9\% & 51.3\% & 65.5\% & 80.3\% & 197.3\% \\
\texttt{glm-5.1} & 12.7\% & 73.1\% & 88.8\% & 92.9\% & 54.5\% \\
\texttt{claude-haiku-4-5} & 8.2\% & 66.1\% & 84.4\% & 95.8\% & 16.4\% \\
\texttt{minimax-m2.5} & 5.0\% & 48.2\% & 70.1\% & 91.8\% & 30.0\% \\
\texttt{grok-4.20} & 8.6\% & 67.9\% & 93.2\% & 82.2\% & 143.2\% \\
\texttt{glm-4.7-flash} & 3.6\% & 32.4\% & 61.7\% & 85.6\% & 76.8\% \\
\texttt{gemini-3.1-flash-lite} & 4.1\% & 34.9\% & 40.4\% & 82.6\% & 95.0\% \\
\texttt{mimo-v2-flash} & 1.8\% & 48.1\% & 78.3\% & 85.5\% & 73.6\% \\
\texttt{gpt-5.4-mini} & 2.7\% & 25.2\% & 58.2\% & 84.6\% & 35.5\% \\
\texttt{gpt-5.4-nano} & 0.9\% & 20.7\% & 48.4\% & 81.6\% & 24.1\% \\
\bottomrule
\end{tabular}
\caption{Full score-consistency metrics for free-form prompt writing over all 33 evaluated models.}
\label{tab:full-score-metrics-prompt-writing}
\end{table}

\section{Additional Topology Details}
\label{app:topology-details}

\begin{figure}[ht]
\centering
\includegraphics[width=0.8\textwidth]{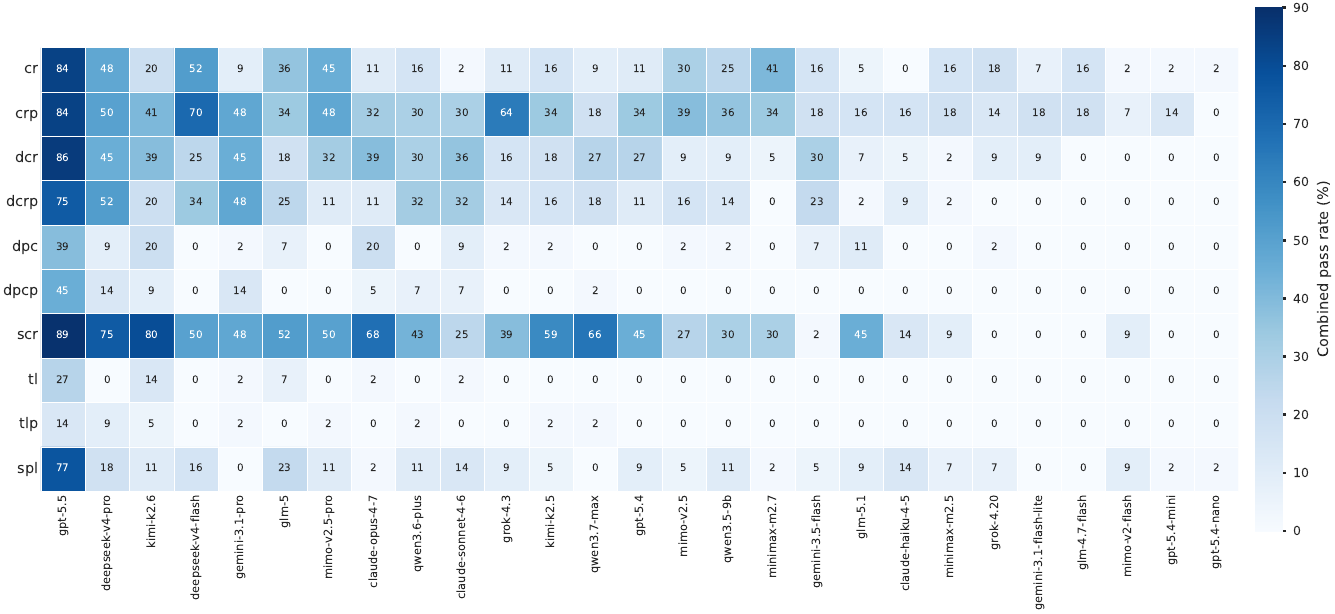}
\caption{Combined pass rate by topology and model. Darker squares indicate higher pass rates.}
\label{fig:topology-model-heatmap}
\end{figure}

Table~\ref{tab:topology-details} lists the topology aliases used in the appendix tables and the number of roles in each template.
Figure~\ref{fig:topology-model-heatmap} reports the combined pass rate for each model and topology.

\begin{table}[ht]
\centering
\small
\begin{tabular}{llr}
\toprule
Alias & Topology & Roles \\
\midrule
\texttt{cr}   & \texttt{coder\_reviewer} & 2 \\
\texttt{crp}  & \texttt{coder\_reviewer\_pool} & 2 \\
\texttt{dcr}  & \texttt{dispatcher\_coder\_reviewer} & 3 \\
\texttt{dcrp} & \texttt{dispatcher\_coder\_reviewer\_pool} & 3 \\
\texttt{dpc}  & \texttt{dispatcher\_planloop\_codeloop} & 5 \\
\texttt{dpcp} & \texttt{dispatcher\_planloop\_codeloop\_pool} & 6 \\
\texttt{scr}  & \texttt{dispatcher\_scientist\_coder\_reviewer} & 4 \\
\texttt{dtc}  & \texttt{dispatcher\_theoryloop\_codeloop} & 5 \\
\texttt{dtcp} & \texttt{dispatcher\_theoryloop\_codeloop\_pool} & 5 \\
\texttt{spl}  & \texttt{supervisor\_phdstudent\_librarian} & 3 \\
\bottomrule
\end{tabular}
\caption{Additional details for the 10 topology templates used in PerspectiveGap.}
\label{tab:topology-details}
\end{table}

\section{Few-Shot Prompting Ablation}
\label{app:fewshot}

\begin{figure}[ht]
\centering
\includegraphics[width=0.6\columnwidth]{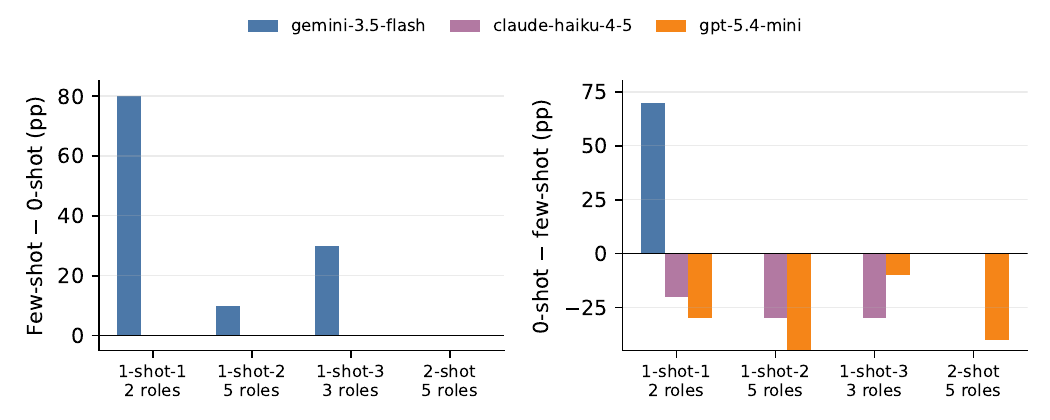}
\caption{Few-shot effects on free-form prompt writing. The left panel shows pass-rate lift, measured as few-shot minus 0-shot. The right panel shows leakage reduction, measured as 0-shot minus few-shot. Higher is better in both panels.}
\label{fig:few-shot-ablation}
\end{figure}

This ablation asks whether few-shot examples improve accuracy on PerspectiveGap's free-form prompt-writing task.
We run it on gpt-5.4-mini, claude-haiku-4-5, and gemini-3.5-flash, where weak 0-shot performance leaves more room for few-shot examples to help.
The comparison must avoid giving away the answer through superficial overlap.
Several hand-written templates share role names or fragment wording, so we manually choose examples whose topology, roles, and content differ from the evaluation slice.
The four settings are listed in Table~\ref{tab:few-shot-settings}; each setting evaluates the ten instances under one target topology.
The main benchmark has 10 topology templates, 100 instances, and 110 scenarios in total once the 10 hand-written templates are included.
The 2-shot setting uses the same 5-role evaluation slice as 1-shot-2, isolating whether a second example helps.

\begin{table}[h]
\centering
\footnotesize
\begin{tabular}{@{}clcc@{}}
\toprule
Setting & Eval. topology & Roles & Example topology \\
\midrule
1-shot-1 & \texttt{cr}  & 2 & \texttt{spl} \\
1-shot-2 & \texttt{dtcp} & 5 & \texttt{cr} \\
1-shot-3 & \texttt{spl} & 3 & \texttt{dpc} \\
2-shot & \texttt{dtcp} & 5 & \texttt{cr}, \texttt{spl} \\
\bottomrule
\end{tabular}
\caption{Few-shot settings used in the ablation. Each evaluation setting uses the ten instances under one target topology. Example topologies were selected manually to avoid shared topology, role names, and fragment wording with the evaluation topology. Topology aliases are listed in Table~\ref{tab:topology-details}.}
\label{tab:few-shot-settings}
\end{table}

Figure~\ref{fig:few-shot-ablation} shows that the worked example helps \texttt{gemini-3.5-flash} in the smaller settings, with +80 pp on 1-shot-1 and +30 pp on 1-shot-3.
The same model gets little or no lift in the 5-role settings.
For \texttt{claude-haiku-4-5} and \texttt{gpt-5.4-mini}, no evaluated output crosses the strict pass threshold in any setting; this does not rule out sub-threshold improvements.
The leakage panel shows why pass-rate lift alone is incomplete: \texttt{gemini-3.5-flash} reduces leakage on 1-shot-1, while \texttt{claude-haiku-4-5} and \texttt{gpt-5.4-mini} leak more under the same intervention.

\section{Distractor-count Ablation}
\label{app:distractor-count}

\begin{figure}[ht]
\centering
\includegraphics[width=0.3\columnwidth]{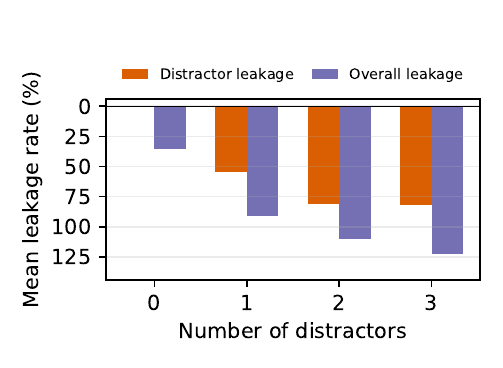}
\caption{Role-fragment assignment leakage as the number of injected distractors increases. Downward bars indicate leakage. Each bar averages over a six-model panel: one flagship and one fast model from OpenAI, Anthropic, and Google.}
\label{fig:leakage-by-distractor-count}
\end{figure}

To study how leakage changes as irrelevant context grows, we rerun role-fragment assignment with 0, 1, 2, or 3 injected distractors on gpt-5.5, gpt-5.4, claude-opus-4-7, claude-sonnet-4-6, gemini-3.1-pro, and gemini-3.5-flash.
We choose this six-model panel to cover flagship and fast models from OpenAI, Anthropic, and Google.
Figure~\ref{fig:leakage-by-distractor-count} plots the model-averaged distractor leakage and overall leakage at each distractor count.
Both rates rise sharply after the first distractor; overall leakage continues to increase through three distractors, reaching 122.3\% on average.

\section{Scratchpad Ablation}
\label{app:scratchpad}

\begin{figure}[ht]
\centering
\includegraphics[width=0.6\columnwidth]{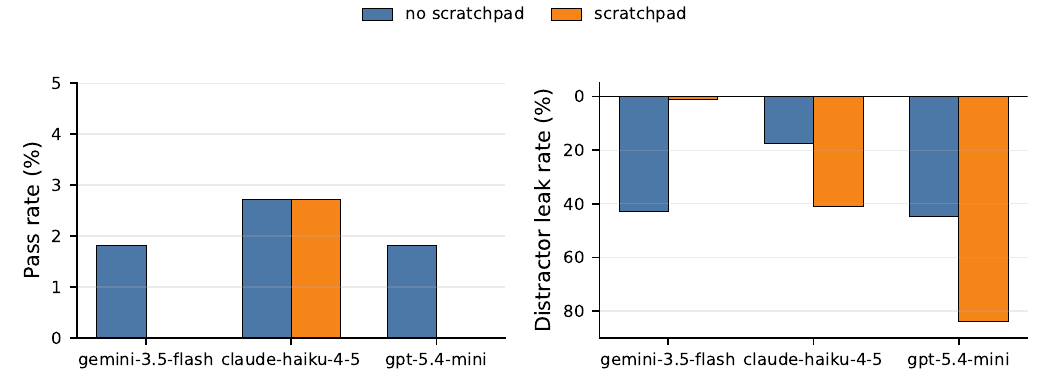}
\caption{Free-form prompt writing pass rate (left) and distractor leak rate (right) on three small models, with and without a hidden scratchpad block. Downward bars in the right panel indicate leakage.}
\label{fig:hidden-scratchpad-ablation}
\end{figure}

Prior work reports that giving models a scratchpad can improve accuracy on self-modeling tasks \citep{ackerman2026selfmodeling}.
We test whether the same trick helps in PerspectiveGap's free-form prompt-writing task.

We use the same model set as Appendix~\ref{app:fewshot}: gpt-5.4-mini, claude-haiku-4-5, and gemini-3.5-flash.
For each model, we rerun the task on all 110 scenarios.
The treatment adds an instruction to first write a hidden scratchpad block listing which fragments each sub-agent needs and why.
The scratchpad block is stripped before scoring, so the scorer checks only the final orchestra text.

Figure~\ref{fig:hidden-scratchpad-ablation} shows that the scratchpad does not improve strict pass rate.
It leaves claude-haiku-4-5 unchanged at 2.7\%, and moves gemini-3.5-flash and gpt-5.4-mini from 1.8\% to 0.0\%.
The leakage effect is model-specific: gemini-3.5-flash drops from 42.7\% to 0.9\%, while claude-haiku-4-5 rises from 17.3\% to 40.9\% and gpt-5.4-mini rises from 44.5\% to 83.6\%.
Scratchpads change how these small models handle distractors, but in this setting they do not solve free-form prompt writing.

\section{Reasoning Effort Ablation}
\label{app:effort}

We ask how free-form prompt writing changes when the same model is given less or more reasoning effort.
We compare gpt-5.5 and claude-sonnet-4-6 at low, medium, and high effort on the same 30-scenario subset used in Appendix~\ref{app:fewshot}.

\begin{figure}[ht]
\centering
\includegraphics[width=0.6\columnwidth]{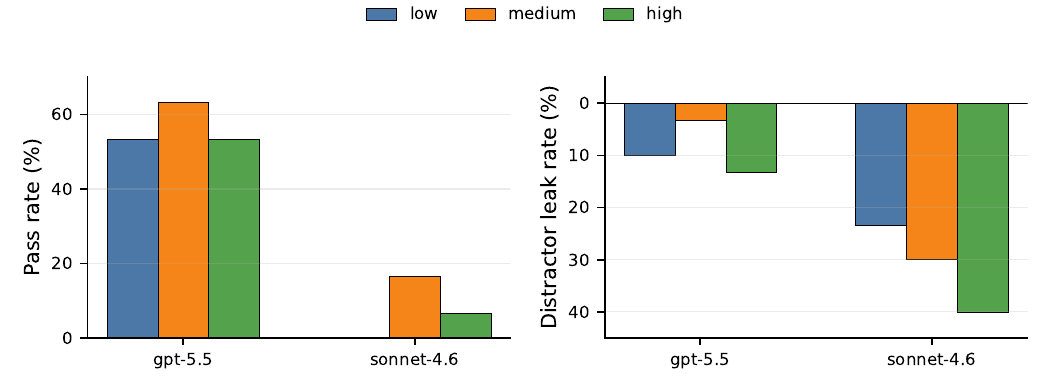}
\caption{Free-form prompt writing pass rate (left) and distractor leak rate (right) across three reasoning-effort levels. Each model has one bar for low, medium, and high effort; downward bars in the right panel indicate leakage.}
\label{fig:reasoning-effort-ablation}
\end{figure}

Figure~\ref{fig:reasoning-effort-ablation} shows medium effort as the pass-rate sweet spot for both models.
gpt-5.5 falls from 63.3\% at medium effort to 53.3\% at both low and high effort.
claude-sonnet-4-6 rises from 0.0\% at low effort to 16.7\% at medium effort, then drops to 6.7\% at high effort.
More effort also makes distractor use worse: gpt-5.5 leaks more at high effort than at medium effort, and claude-sonnet-4-6 leakage rises from 23.3\% to 40.0\% across the sweep.

\section{Scorer Agreement with Human Labels}
\label{app:agreement}

For the free-form prompt-writing task, the scorer test set contains 716 rows sampled from benchmark outputs and labeled by human annotators for fragment containment.
It is used only to validate the rule scorer; the scorer itself reads only the benchmark reference mapping at evaluation time and does not consult any human labels.

\begin{table}[ht]
\centering
\small
\begin{tabular}{lr}
\toprule
Metric & Value \\
\midrule
Percent agreement & 99.44\% (712/716) \\
F1 (pass class)   & 99.57\% \\
Precision (pass)  & 99.78\% \\
Recall (pass)     & 99.36\% \\
\bottomrule
\end{tabular}
\caption{Scorer agreement with human labels on all 716 free-form prompt writing test-set rows.}
\label{tab:scorer-agreement}
\end{table}

\section{Role-Fragment Assignment Trivial Baselines}
\label{app:baselines}

To validate that the leaderboard is not driven by surface shortcuts, we analytically score three trivial baselines on the same role-fragment assignment set-equality scorer used in the main results.
Table~\ref{tab:trivial-baselines} reports the results.
All three baselines reach a 0.0\% pass rate, so the nonzero leaderboard scores cannot be explained by copy-all, role-name keyword matching, or random assignment.

\begin{table}[ht]
\centering
\small
\begin{tabular}{@{}p{0.64\linewidth}r@{}}
\toprule
Baseline & Role-fragment assignment pass rate \\
\midrule
Copy-all (every role gets every fragment) & 0.0\% \\
Role-keyword (assign fragment to roles whose name token appears in its heading) & 0.0\% \\
Random (each fragment to one uniform-random role, 20 random draws) & 0.0\% \\
\bottomrule
\end{tabular}
\caption{Three trivial baselines on the role-fragment assignment set-equality scorer.}
\label{tab:trivial-baselines}
\end{table}

\section{Professional-Domain Coverage}
\label{app:professional-domains}

Table~\ref{tab:professional-domain-coverage} summarizes the domains used in PerspectiveGap.

\begin{table*}[!htbp]
\centering
\scriptsize
\setlength{\tabcolsep}{4pt}
\begin{tabular}{p{0.36\linewidth}rp{0.50\linewidth}}
\toprule
Professional domain & Scenarios & Example role list (one scenario per domain) \\
\midrule
Medical, clinical trials, regulatory safety  & 14 & care coordinator, diagnostician, diagnostic challenge physician, treatment planner, patient-safety reviewer \\
Litigation, criminal defense, appellate      & 13 & docket coordinator, appellate strategist, red-team appellate counsel, brief drafter, opposing-counsel brief reviewer \\
Cultural heritage, provenance, authentication & 11 & authentication case manager, provenance researcher, provenance skeptic, conservation science examiner, forensic methods auditor \\
Intelligence, security operations, vulnerability research & 10 & secops director, threat modeler, security architect, red-teamer, vulnerability analyst \\
Tax law, IRS audit, forensic accounting       & 7  & audit docket clerk, tax credit preparer, IRS examining agent \\
Civil and structural engineering, plan check  & 7  & project coordinator, structural scheme designer, design revision engineer, code-compliance examiner, calculation package engineer, independent structural reviewer \\
Investigative journalism, editorial           & 7  & assigning editor, investigative reporter, skeptical investigations editor, story writer, fact-checking editor \\
Finance, quant trading, M\&A, risk            & 6  & deal orchestrator, valuation strategist, investment committee skeptic, term-sheet drafter, due-diligence auditor \\
Drug discovery, pharmacology, toxicology      & 4  & program director, pharmacologist, toxicology skeptic, trial designer, regulatory auditor \\
Environmental permitting, impact assessment   & 4  & permitting coordinator, impact assessment lead, regulatory challenge reviewer, EIA report author, permit-compliance auditor \\
Aerospace mission operations, flight dynamics & 3  & mission commander, astrodynamics architect, maneuver adjuster, orbital-mechanics reviewer, burn-sequence compiler, physics-simulator proxy \\
Patent law and IP                             & 3  & patent attorney, patent examiner \\
Policy, campaign, legislation                 & 3  & policy coordinator, macro-theorist, intervention tuner, economic skeptic, legislation drafter, compliance stress-tester \\
Other (game and screen production, fire incident response, mathematical proof verification, anomaly detection, culinary operations) & 8 & showrunner, narrative architect, plot-hole critic, scene writer, continuity auditor \\
Software engineering and research & 10 & coder, reviewer, dispatcher, scientist, theorist, phdstudent, librarian \\
\bottomrule
\end{tabular}
\caption{Professional-domain coverage of PerspectiveGap.}
\label{tab:professional-domain-coverage}
\end{table*}

\section{Concrete Reference-Mapping Example}
\label{app:reference-mapping-example}

This appendix makes the reference policy auditable on an actual benchmark scenario.
We use the \texttt{dispatcher\_theoryloop\_codeloop} scenario.
It shows the background given to the model, the fragment headings, the \emph{``need-only''} instruction, the reference assignment, and textual evidence for representative boundary decisions.

\begin{quote}
\begin{tcolorbox}[colback=gray!6,colframe=gray!35,boxrule=0.35pt,arc=1pt,left=6pt,right=6pt,top=5pt,bottom=5pt,breakable]
\small
I need you to set up a 5-agent pipeline that solves mathematical optimization problems by developing theoretical analysis and solver code in parallel, with each side informing the other. A problem arrives as a natural-language statement in \texttt{problem.md}; read it before starting your part.

The five agents:
\begin{itemize}[leftmargin=*]
\item \textbf{dispatcher} orchestrates the run by alternating one theory step with one code step and tracking progress. It does no modeling, coding, or reviewing itself.
\item \textbf{theory loop} --- the \textbf{theorist} writes the theoretical analysis to \texttt{THEORY.md}; the \textbf{theory-reviewer} judges it. The theorist reads \texttt{SOLUTION.md} on each revision pass to incorporate empirical evidence.
\item \textbf{code loop} --- the \textbf{coder} writes solver code and its results to \texttt{SOLUTION.md}; the \textbf{code-reviewer} judges them. The coder reads \texttt{THEORY.md} on each revision pass to align the implementation with the latest theoretical claims.
\end{itemize}

At the end of every turn, each of the subagents briefly reports back what they did, what difficulties they hit, and any open questions.

The pipeline hands work off through three files:
\begin{itemize}[leftmargin=*]
\item \texttt{problem.md} --- the original problem, read-only. All four domain agents read it.
\item \texttt{THEORY.md} --- the theoretical analysis. Theorist writes; theory-reviewer reviews; coder and code-reviewer read it to know which bounds and invariants the implementation must respect.
\item \texttt{SOLUTION.md} --- the solver code and its results. Coder writes; code-reviewer reviews; theorist and theory-reviewer read it to ground the theory in observed numbers.
\end{itemize}

Each reviewer records its score as a \texttt{<review score=X>} block inside the file it reviewed, overwriting any previous block. The dispatcher reads those scores to drive the loops.

Unlike a plan-then-code pipeline where the plan must converge before the code starts, here neither side ever finishes first. Both loops alternate forever, capped at a fixed iteration count. Theory may tighten after seeing code's empirical numbers; code may correct after seeing theory's refined bounds.
\end{tcolorbox}
\end{quote}

The prompt then shows the fragment headings in Table~\ref{tab:reference-mapping-fragments}.
We omit the fragment bodies here, but keep the headings because they show how the scenario separates actor instructions, reviewer instructions, orchestration, and reporting.

\begin{table}[!htb]
\centering
\small
\begin{tabular}{@{}ll@{}}
\toprule
Fragment & Heading \\
\midrule
\texttt{f1} & problem context \\
\texttt{f2} & theorist deliverable rule \\
\texttt{f3} & theory toolbox \\
\texttt{f4} & coder deliverable rule \\
\texttt{f5} & solution-doc writing rule \\
\texttt{f6} & theory-review rule \\
\texttt{f7} & code-review rule \\
\texttt{f8} & reviewer scoring rule \\
\texttt{f9} & dispatch rule \\
\texttt{f10} & theorist workflow \\
\texttt{f11} & coder workflow \\
\texttt{f12} & report-back instruction \\
\bottomrule
\end{tabular}
\caption{Fragment headings for the \texttt{dispatcher\_theoryloop\_codeloop} example.}
\label{tab:reference-mapping-fragments}
\end{table}

After the fragment block, the prompt tells the model: \emph{``Each agent's prompt should contain only the information that agent needs to do its job.''}
and the output format.
Table~\ref{tab:reference-mapping-example} shows the resulting reference assignment.

\begin{table}[!htb]
\centering
\small
\begin{tabular}{@{}p{0.18\linewidth}p{0.24\linewidth}p{0.52\linewidth}@{}}
\toprule
Role & Reference fragments & Rationale \\
\midrule
Dispatcher & \texttt{f9} & The background says the dispatcher \emph{``orchestrates the run by alternating one theory step with one code step and tracking progress''} and \emph{``does no modeling, coding, or reviewing itself.''} It therefore needs the dispatch rule only. \\
Theorist & \texttt{f1}, \texttt{f2}, \texttt{f3}, \texttt{f10}, \texttt{f12} & The theorist needs the problem context, its deliverable rule, the theory toolbox, its workflow across turns, and the instruction to report back. \\
Theory-reviewer & \texttt{f1}, \texttt{f3}, \texttt{f6}, \texttt{f8}, \texttt{f12} & The theory-reviewer needs the problem context, enough theory-tool context to judge the analysis, the theory-review rule, the reviewer scoring rule, and the instruction to report back. \\
Coder & \texttt{f1}, \texttt{f4}, \texttt{f5}, \texttt{f11}, \texttt{f12} & The coder needs the problem context, its deliverable rule, solution-document writing rules, its workflow across turns, and the instruction to report back. \\
Code-reviewer & \texttt{f1}, \texttt{f7}, \texttt{f8}, \texttt{f12} & The code-reviewer needs the problem context, the code-review rule, the reviewer scoring rule, and the instruction to report back. \\
\bottomrule
\end{tabular}
\caption{Reference role-fragment assignment for one PerspectiveGap scenario.}
\label{tab:reference-mapping-example}
\end{table}

The main boundary decisions in this example are not arbitrary.
They follow from the stated role responsibilities:
\begin{enumerate}[leftmargin=*]
\item \texttt{f12} belongs to all four domain agents because the background says \emph{``At the end of every turn, each of the subagents briefly reports back what they did, what difficulties they hit, and any open questions.''} It does not belong to the dispatcher: the dispatcher receives those reports, but \texttt{f12} tells domain agents what to report.
\item \texttt{f6} belongs to the theory-reviewer, not to the theorist. Its first sentence is \emph{``Review THEORY.md against problem.md and against the numbers reported in SOLUTION.md (if any).''} Giving it to the theorist leaks the critic's instructions into the actor's prompt.
\item \texttt{f2} belongs to the theorist, not to the theory-reviewer. It begins with \emph{``Produce the theoretical analysis for the problem and write it to THEORY.md.''} Although it is about theory, it specifies the actor's deliverable rather than the reviewer's job.
\item The code side follows the same boundary. \texttt{f4} begins with \emph{``Read problem.md and THEORY.md, then implement the solver:''}; \texttt{f5} says \emph{``Keep SOLUTION.md lean.''}; and \texttt{f11} says \emph{``The solver code and its results live in SOLUTION.md, and you may be invoked on it more than once.''} These are coder-side instructions. By contrast, \texttt{f7} begins with \emph{``Review SOLUTION.md against problem.md and against THEORY.md,''} and \texttt{f8} says to \emph{``Score the work from 1 to 10''}; those are reviewer-side instructions.
\end{enumerate}

\end{document}